\documentclass[letterpaper, 10 pt, conference]{ieeeconf}  

\usepackage{amssymb}
\usepackage{graphicx}
\usepackage{amsmath}

\usepackage{booktabs}
\usepackage{caption}
\usepackage{tikz}

\usepackage{bm}
\usepackage{bbm}
\usepackage{mathtools}

\usepackage{textcomp}
\usepackage{xcolor}
\usepackage{wrapfig}
\usepackage{cite}
\usepackage{algorithm}
\usepackage{algorithmic}
\usepackage{dsfont}
\usepackage{multirow}
\usepackage{tabularx}
\usepackage{colortbl} 
\usepackage[dvipsnames]{xcolor} 

\usepackage{subcaption}
\usepackage{hyperref}

\usepackage[capitalize]{cleveref}
\crefname{section}{Sec.}{Secs.}
\Crefname{section}{Section}{Sections}
\Crefname{table}{Table}{Tables}
\crefname{table}{Tab.}{Tabs.}

\def \poses{\mathcal{X}}
\def \landmarks{\mathcal{L}}

\def \measurements{\mathcal{Z}}

\newcommand{\X}[1]{\mathbf{X}_{#1}}      
\newcommand{\Xext}[1]{\tilde{\mathbf{X}}^{\mathrm{3DGS}}_{#1}} 
\newcommand{\Logmap}{\operatorname{Log}}     

\newcommand{\wnorm}[2]{\left\| #1 \right\|^2_{#2}} 

\newcommand{\Iext}{\mathcal{I}_{\mathrm{3DGS}}}

\newcommand{\Sigext}[1]{\Sigma^{\mathrm{3DGS}}_{#1}}

\newcommand{\Method}{ReefMapGS}

\newcommand{\loss}{\mathcal{L}}     

\newcommand{\link}[1]{\textcolor{magenta}{\href{#1}{#1}}}

\colorlet{colorFst}{Green!25}       
\colorlet{colorSnd}{SpringGreen!45} 
\colorlet{colorTrd}{Yellow!30}      
\colorlet{colorLow}{darkgray!30}    
\newcommand{\fst}{\cellcolor{colorFst}}   
\newcommand{\nd}{\cellcolor{colorSnd}}      
\newcommand{\rd}{\cellcolor{colorTrd}}      

\newcommand{\cg}{\color{gray!80}}

\IEEEoverridecommandlockouts                              

\title{\LARGE \bf
ReefMapGS: Enabling Large-Scale Underwater Reconstruction by Closing the Loop Between Multimodal SLAM and Gaussian Splatting
}
\author{Daniel Yang$^{1, 2}$, Jungseok Hong$^{1}$, John J. Leonard$^{1}$, and Yogesh Girdhar$^{2}$
\thanks{\scriptsize $^{1}$Massachusetts Institute of Technology $^{2}$Woods Hole Oceanographic Institution}%
\thanks{}%
}

\makeatletter
\let\oldtwocolumn\twocolumn
\renewcommand\twocolumn[1][]{%
  \oldtwocolumn[{#1%
    \centering
    \vspace{-1.0em}
    \includegraphics[width=\textwidth]{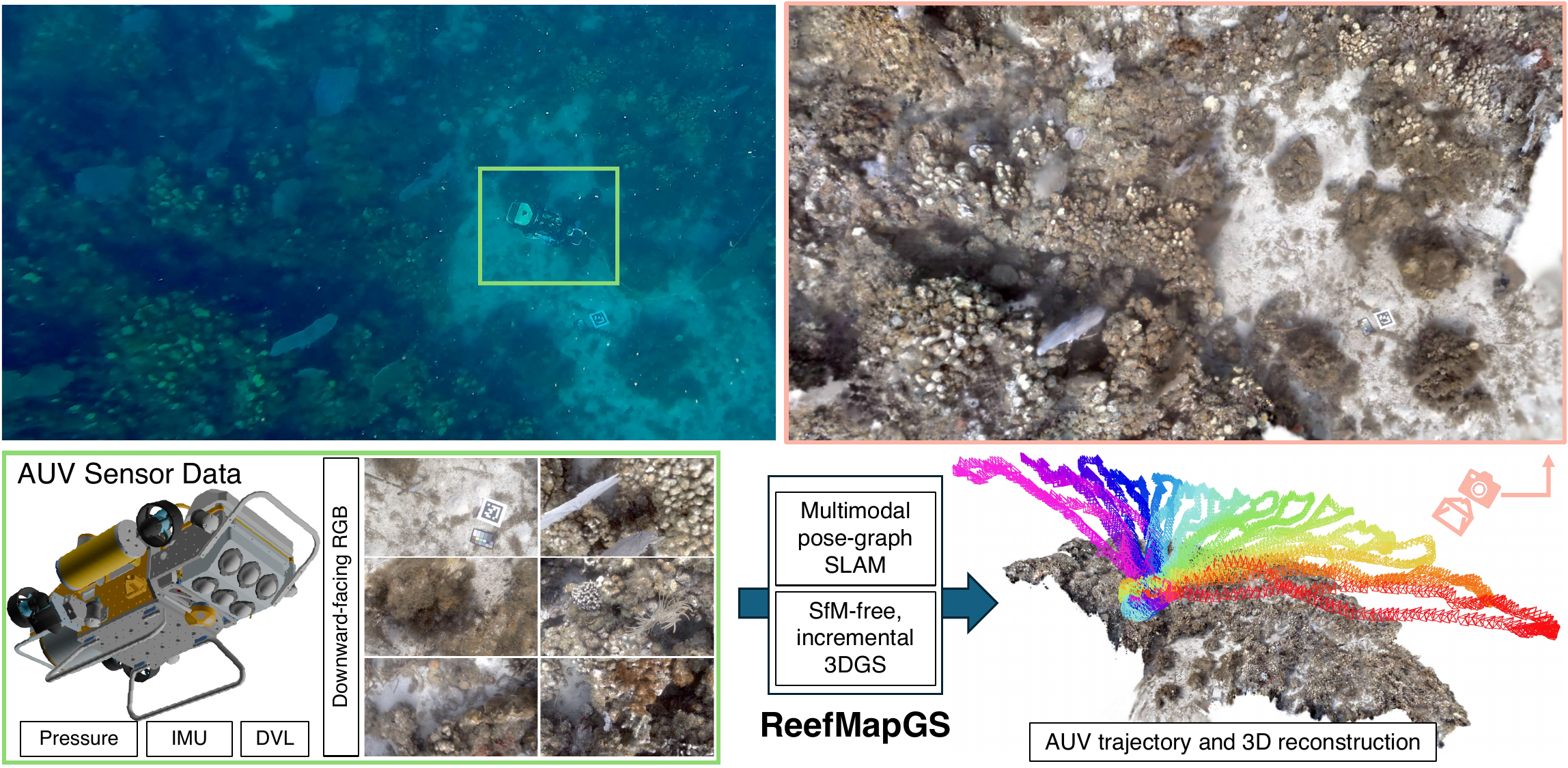}
    \vspace{-1.0em}
    \captionsetup{type=figure}
    \captionof{figure}{%
      \textbf{\Method} integrates multimodal pose-graph SLAM with 3D Gaussian Splatting to enable rapid, dense reconstruction of underwater environments like coral reefs without needing computationally expensive structure-from-motion pipelines. On the top left, we show an image captured by a snorkeler on the water surface passively observing a robot performing a visual benthic survey of a coral reef. Combining the AUV-collected monocular imagery with other vehicle sensor data, \Method can rapidly estimate the whole robot trajectory while also providing a high-quality, dense 3DGS reconstruction of the whole scene. The top right view is synthetically rendered from the obtained 3DGS reconstruction.
    }%
    \label{fig:intro}
  }]%
}
\makeatother

\begin{document}
\maketitle

\begin{abstract}
3D Gaussian Splatting is a powerful visual representation, providing high-quality and efficient 3D scene reconstruction, but it is crucially dependent on accurate camera poses typically obtained from computationally intensive processes like structure-from-motion that are unsuitable for field robot applications. However, in these domains, multimodal sensor data from acoustic, inertial, pressure, and visual sensors are available and suitable for pose-graph optimization-based SLAM methods that can estimate the vehicle’s trajectory and thus our needed camera poses while providing uncertainty. We propose a 3DGS-based incremental reconstruction framework, \Method, that builds an initial model from a high certainty region and progressively expands to incorporate the whole scene. We reconstruct the scene incrementally by interleaving local tracking of new image observations with optimization of the underlying 3DGS scene. These refined poses are integrated back into the pose-graph to globally optimize the whole trajectory. We show COLMAP-free 3D reconstruction of two underwater reef sites with complex geometry as well as more accurate global pose estimation of our AUV over survey trajectories spanning up to 700 m. Project page: \link{https://reefmapgs.github.io/}.
\end{abstract}
\vspace{-0.5em}

\section{Introduction}
\label{sec:intro}

Recent advances in autonomous underwater vehicles (AUVs) and underwater mapping have enabled the mapping of challenging underwater environments, an essential capability for environmental exploration and monitoring \cite{girdhar_curee_2023, joshi2022underwater, johnson2017high, merveille2024advancements, williams_simultaneous_2004}. For many scientific and ecological tasks, such as coral reef monitoring and large-scale spatiotemporal environmental surveys, we would like to rapidly obtain dense, high-quality scene representations. The quality of dense reconstruction depends on the input images and the accuracy of camera pose estimation. However, obtaining these inputs for underwater scenes remains challenging due to degraded visual conditions (e.g., light attenuation, turbidity, color distortion, and low-texture scenes) and limited sensing capabilities, resulting in persistent inaccuracies in pose estimation.

To estimate poses in batch, offline structure-from-motion (SfM) and multi-view stereo (MVS) methods have been used, but they require substantial computational resources. Visual simultaneous localization and mapping (SLAM) methods can be used to estimate poses incrementally and build maps in real time, but they are fundamentally unreliable underwater due to the challenges mentioned above. To address this issue, state-of-the-art underwater SLAM systems integrate multimodal sensors, combining inertial, velocity, and depth measurements with visual and/or acoustic cues in pose-graph optimization frameworks. While these systems effectively estimate vehicle trajectories over long distances, the resulting maps are typically sparse and optimized for navigation rather than dense scene reconstruction. Consequently, existing systems are generally inadequate for downstream scientific analysis that requires detailed information about the target environments. This fundamental mismatch between existing systems and scalable, dense reconstruction remains a key bottleneck for using such maps across a wide range of tasks.

Recently, 3D Gaussian Splatting (3DGS) \cite{kerbl_3d_2023} has emerged as an efficient and differentiable representation that produces high-fidelity, dense reconstructions while allowing for pose refinement via gradient-based optimization. This advancement has motivated the development of 3DGS-based SLAM systems \cite{keetha_splatam_2024, zheng_wildgs-slam_2025, matsuki_gaussian_2024}. However, these systems are primarily designed for terrestrial environments and rely on reliable visual characteristics or RGB-D sensing, assumptions that do not hold underwater, where sensor measurements are noisier.

We address this gap by tightly coupling multimodal pose-graph SLAM with incremental updates to a 3DGS model. Rather than treating dense reconstruction as a post-processing step or relying on batch-based SfM, we propose a closed loop between pose estimation from SLAM and scene representation. Starting from regions of low pose uncertainty near a known landmark, we incrementally expand the dense scene representation. As the 3DGS model improves, it is used to refine camera poses via differentiable rendering, and these refined poses are added to the pose graph to improve global trajectory estimation. 

This bidirectional refinement allows pose estimation and dense reconstruction to mutually improve each other over extended trajectories, even when initial pose estimates are noisy, input images are distorted, and visual overlap across the scene is minimal. In contrast to previous 3DGS-based SLAM systems, our approach leverages a known landmark used in foveated, rosette-shaped surveys performed by AUVs in coral reef benthic surveys as well as the complementary strengths of multimodal SLAM and differentiable scene representations. The resulting pipeline is bundle-adjustment-free and capable of dense, incremental reconstruction at scales relevant to real-world underwater surveys, spanning hundreds of meters while also improving global pose accuracy.

Our contributions are as follows:
\begin{enumerate}
    \item A novel framework for building large-scale reconstructions by integrating 3D Gaussian Splatting with pose-graph SLAM, achieving both increased 3D reconstruction quality and less error in robot trajectory estimation, suitable for field robotics applications.
    \item A dataset consisting of multimodal robot sensor data from two real-world coral reef surveys at biologically interesting sites with complex geometry, spanning 347 m over 20 m x 10 m and 696 m over 20 m x 20 m.
\end{enumerate}

\section{Related works}
\label{sec:related}

\subsection{Bundle adjustment-free 3D reconstruction}
Recent feed-forward 3D models reduce reliance on classical SfM by predicting point maps directly from pairs of images, from which camera parameters can be estimated.
DUSt3R \cite{wang_dust3r_2024} introduced a framework for dense per-pixel 3D pointmap regression, enabling end-to-end reconstruction from arbitrary image pairs without camera calibrations. Many works build upon DUSt3R to improve dense feature matching \cite{cabon2025must3r} and introduce strategies for handling multiple views \cite{duisterhof_mast3r-sfm_2024, wang2025cut3r} or a stream of images \cite{wang2024spann3r, wang2025cut3r}. Other works like VGGT \cite{wang_vggt_2025} utilize feed-forward transformers to simultaneously attend to and jointly reason over all input views. These models rely on learned geometric priors and dense correspondences that can initialize or regularize poses. However, when texture is scarce and viewpoints are suboptimal (e.g., limited field of view, minimal overlap), as is common in underwater scenes, they tend to struggle.

\subsection{Underwater SLAM}
Underwater SLAM is substantially more challenging than terrestrial or aerial SLAM due to degraded visual conditions, limited sensing bandwidth, and the absence of GPS \cite{kinsey2006survey,leonard2016autonomous}. Optical imagery suffers from attenuation, color distortion, turbidity, and low-texture scenes, causing vision-only and visual-inertial SLAM systems to fail or drift even under moderate underwater conditions \cite{joshi2019experimental}. To achieve robust long-horizon navigation, underwater SLAM systems therefore rely on multi-modal sensor fusion, commonly integrating IMU, DVL, and depth measurements within factor-graph or pose-graph optimization frameworks \cite{merveille2024advancements}. Visual or acoustic sensing is typically used to detect features and provide relative pose constraints.
Underwater SLAM systems such as~\cite{rahman2019svin2} focus on robust trajectory estimation using sparse map representations (e.g., keyframes). In contrast, recent works including~\cite{wang2023real, xu2025aqua, song_turtlmap_2024} yield dense maps, but typically evaluate on relatively short trajectories rather than long-horizon mapping and rely on stereo camera inputs.

\subsection{Dense SLAM}
Dense underwater reconstruction has been dominated by SfM and MVS pipelines like COLMAP \cite{schoenberger2016sfm} or commercially Metashape \cite{noauthor_agisoft_nodate}. They provide high-quality models but require computationally intensive bundle adjustment unsuitable for rapid field deployment. Camera pose estimation remains a key bottleneck for scaling dense reconstruction to large underwater environments. One method \cite{sauder_scalable_2024} eschews these pipelines for learning-based ego-motion estimation \cite{bian_unsupervisedscale_2019} to build underwater scene maps, but is limited to simpler transect trajectories and is limited in the fidelity of its output. 

Recent advances in dense scene representations, particularly radiance field methods like NeRF \cite{mildenhall2020nerf} and 3D Gaussian Splatting (3DGS) \cite{kerbl_3d_2023}, offer new opportunities to bridge the gap between sparse SLAM and dense reconstruction. 3DGS \cite{kerbl_3d_2023} has emerged as a computationally efficient explicit representation capable of producing high-fidelity dense geometry and appearance while supporting differentiable optimization. Several recent SLAM systems \cite{keetha_splatam_2024, zheng_wildgs-slam_2025, matsuki_gaussian_2024} incorporate 3DGS to jointly optimize camera poses and dense scene models, demonstrating photo-realistic reconstruction and camera tracking in terrestrial environments, typically controlled spaces such as homes and offices. However, these methods typically assume well-calibrated, illumination-consistent, high-quality inputs with significant overlapping visibility between large portions of the RGB or RGBD input stream. In contrast, benthic imagery from a low-cost AUV may not have these characteristics, with visual degradation from imaging through water and imprecise calibration imaging through a dome port. To overcome these difficulties, this work leverages multimodal sensor data from the AUV and principally integrates pose-graph optimization with dense 3DGS-based scene reconstruction, enabling bundle-adjustment-free large-scale underwater reconstruction while achieving reasonable global vehicle pose estimates.

\section{Method}
\label{sec:method}
\begin{figure*}[ht]
        \centering
        \includegraphics[width=\linewidth]{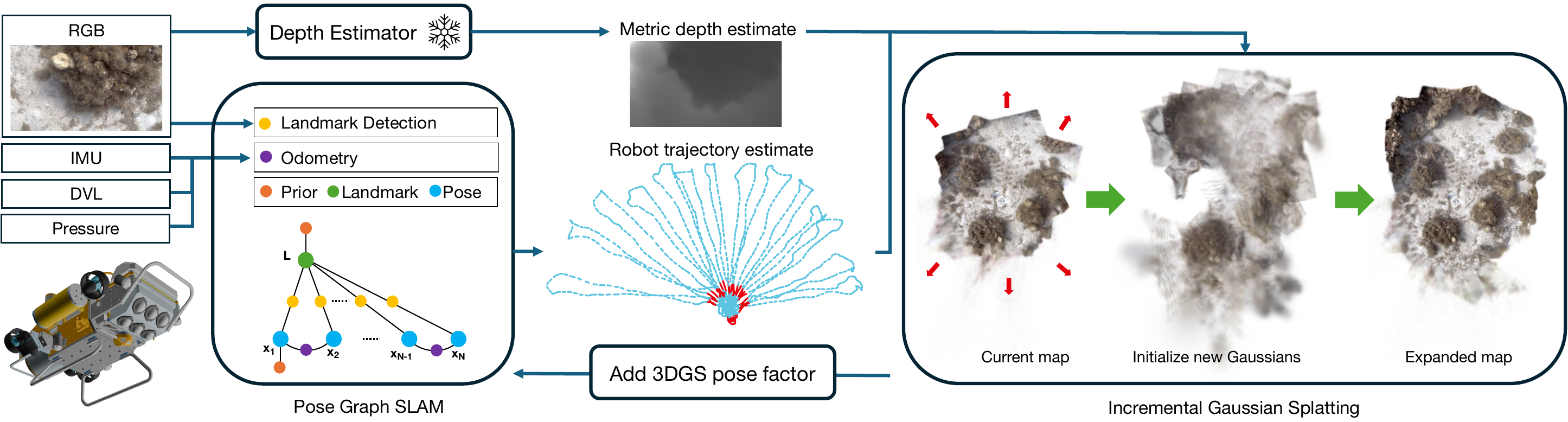}
        \caption{\textbf{System overview} \Method takes as input RGB images and inertial, acoustic, and pressure sensor data from an AUV. Sensor data is fused into odometry while the known, central landmark is detected from RGB frames. Together, this information is integrated into a factor graph to estimate the whole AUV trajectory, shown in the center. We incrementally build a 3DGS model of the scene, starting from the region of highest certainty near the center. The solid blue line shows observations incorporated into the current map. From this current map, we expand and incorporate new observations at the frontier, the solid red line. The observations are aligned with the current map starting from the pose estimate from the factor graph. Together with depth information from a metric depth estimation model, we initialize new Gaussians and then optimize the model, yielding an expanded map. We repeat this process to incorporate the remaining observations, shown in the dashed blue line.}
        \label{fig:overview}
\end{figure*}

\subsection{Problem formulation}

We aim to construct a dense 3D representation of an underwater environment while estimating robot, and thus camera, poses through fusing multimodal sensor data. We assume that a robot performs a survey of an underwater scene centered around a fixed, known landmark. While here we consider a robot navigating along a rosette trajectory, a flower-petal pattern as shown in \cref{fig:2d-trajs}, \Method is applicable to other survey trajectories centered around a fixed, known landmark.

Let $\mathcal{X} = \{\mathbf{X}_k\}^{K}_{k=0}$ denote the discrete sequence of robot camera poses, where each pose $\mathbf{X}_k \in SE(3)$. $\mathcal{Z}$ denotes the sensor measurements which arrive asynchronously at sensor-specific timestamps: Monocular RGB images $\mathcal{I} = \{(t_i, \mathbf{I}_{t_i})\}_{i=1}^{N_{\text{rgb}}}$ from a downward-facing calibrated camera, linear velocities $\mathcal{V} = \{(t_j, \mathbf{v}_{t_j})\}_{j=1}^{N_{\text{dvl}}}$, $\mathbf{v}_{t_j} \in \mathbb{R}^{3}$ from a DVL, angular velocity $\boldsymbol{\Omega} = \{(t_j, \boldsymbol{\omega}_{t_j})\}_{j=1}^{N_{\text{imu}}}$, $\boldsymbol{\omega}_{t_j} \in \mathbb{R}^{3}$ and linear acceleration $\mathcal{A} = \{(t_j, \mathbf{a}_{t_j})\}_{j=1}^{N_{\text{imu}}}$, $\mathbf{a}_{t_j} \in \mathbb{R}^{3}$ from an IMU, and depth measurements $\mathcal{D} = \{(t_\ell, d_{t_\ell})\}_{\ell=1}^{N_{\text{depth}}}$, $d_{t_\ell} \in \mathbb{R}$ from a pressure sensor. Note that depth here refers to distance from the sea surface, not depth from the camera sensor.

\subsection{Pose-graph optimization for trajectory estimation}
We formulate the pose estimation problem as a factor graph optimization problem, where we estimate the robot's camera pose $\mathcal{X}$ and landmark $\mathcal{L}$ via Maximum a Posteriori (MAP) inference given sensor measurements $\mathcal{Z}$.

\vspace{-0.3cm}
\begin{equation}\label{eq:fg}
\mathcal{X}^{\star}, \mathcal{L}^{\star}
=\underset{\poses, \landmarks \in SE(3)}{\arg \max } \; p(\poses, \landmarks \mid \measurements).
\end{equation}
\vspace{-0.3cm}

Rather than creating separate factors for each sensor measurement, which does not scale to longer time horizons, especially with high-frequency sensor data, we first estimate odometry using an Extended Kalman Filter (EKF). We also pre-process the raw IMU measurements, which are typically affected by noise and sensor bias, with a complementary filter \cite{valenti2015keeping} to provide a stable real-time orientation estimate. We fuse the filtered IMU data, depth, and DVL velocity measurements with an EKF \cite{moore2016generalized} to obtain odometry data.

To build a factor graph, we incorporate prior knowledge of a static landmark located at the center of each survey site. We add both odometry and landmark measurement factors into the graph, each associated with robust noise models (e.g., Huber) to handle potential outliers. The final pose trajectory is estimated by solving the factor graph using Levenberg–Marquardt (LM) optimization.

\subsection{Incremental 3D Gaussian Splatting}
\label{subsec:incremental}

\begin{figure*}[ht]
        \centering
        \includegraphics[width=\linewidth]{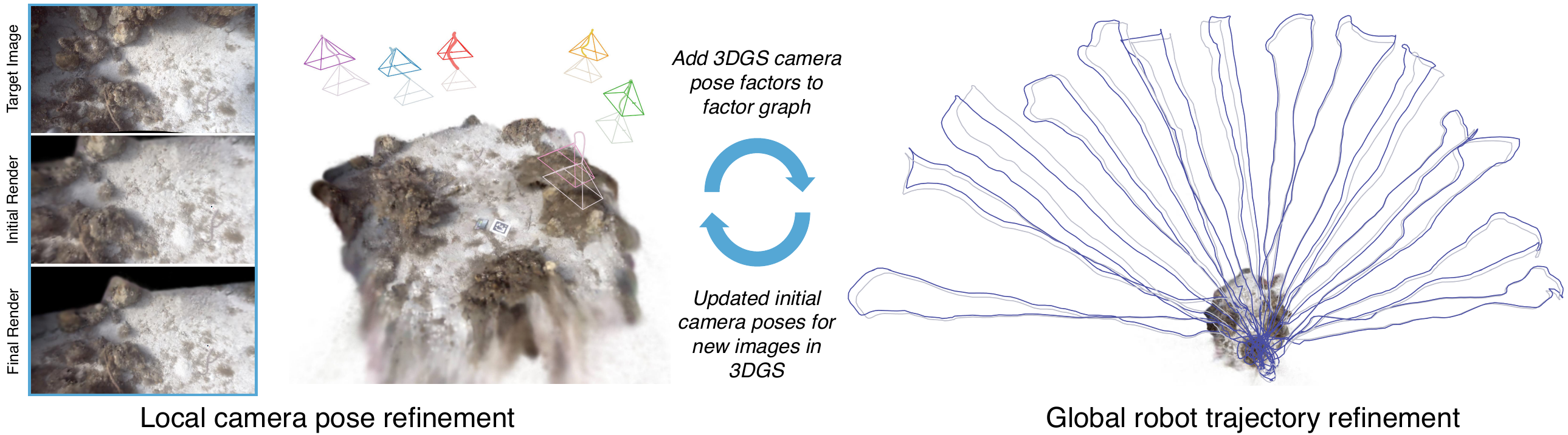}
        \caption{\textbf{Incremental, radially-expanding 3DGS optimization framework.} Local camera pose refinement occurs as we radially and incrementally expand our 3DGS-based map representation, incorporating new cameras and their visual data into the scene. Given a current 3DGS map, target image, and initial camera pose estimate, we can refine the camera pose with gradient descent (as shown from the light to dark frustums). Sample target, initial pose render, and final pose render images are shown on the left for the blue frustum. As camera poses are refined, we add these refined camera pose estimates as factors into our pose-graph to globally update the whole robot trajectory, as shown between the light (before) and dark (after) overlayed trajectories. Local pose refinement aligns new observations before expanding the scene, while global trajectory optimization propagates these changes to poses far away.}
        \label{fig:pose_refinement}
\end{figure*}


\Method utilizes a 3D Gaussian Splatting (3DGS) map representation, parameterizing the scene with isotropic 3D Gaussians and using zero-order spherical harmonics. Thus, each 3D Gaussian has a set of 8 parameters, consisting of the mean $\bm{\mu} \in \mathbb{R}^3$, scale $\bm{S} \in \mathbb{R}^1$, opacity $\bm{o} \in \mathbb{R}^1$, and color $\bm{c} \in \mathbb{R}^3$. This reduced set of parameters requires less memory and simplifies the optimization problem. Typically, a whole 3DGS model is optimized given a dataset of RGB images, corresponding poses, and a point cloud to initialize the 3DGS model, the latter two of which are typically obtained from SfM (e.g., COLMAP \cite{schoenberger2016sfm}). All views of the scene are jointly optimized and needed at the start of optimization.

In contrast to this paradigm of easily available accurate camera poses and a point cloud to initialize the full scene, our underwater, field robotics context has noisier and imprecisely estimated data. Camera poses are estimated by fusing multimodal sensor data while RGB data is combined with a monocular depth estimator to generate point clouds to initialize 3D Gaussians. Additionally, incorporating a fixed landmark into the pose-graph results in spatially varying uncertainty, where poses near the landmark are tightly constrained, while those farther away show much higher uncertainty.

Thus, we propose leveraging the fixed, known landmark to incrementally build the 3DGS scene, starting from the region of least noise, near the central landmark, and progressively introducing and optimizing new observations on the frontier of what has already been incorporated, as shown in the right of \cref{fig:overview}. Concretely, we discretize the robot trajectory based on distance from the central landmark into rings of paired poses and image observations and incrementally integrate each ring into the 3DGS scene. Notably, before incorporating new observations, we locally refine their pose estimates by fitting the observation against the scene optimized thus far, as shown on the left of \cref{fig:pose_refinement}. Given a 3DGS model of the partial scene, a new image observation, and an estimate of the corresponding camera pose, we can refine the camera pose by minimizing reconstruction error and propagating the gradients through the 3DGS model back to the pose as implemented in libraries like gsplat\cite{ye2025gsplat}. This requires visual overlap between the observed and pre-refinement pose rendered image and can fail if the pose estimate is too far away from the actual pose. As the 3DGS model is refined through typical photometric reconstruction and adaptive density control processes, we interleave local pose refinement. Synergistically, improvements in scene quality lead to more accurate pose estimation and vice versa.

However, this local pose refinement process only improves individual camera pose estimates. A minor translational and rotational error for state estimation in a region of low uncertainty could have a large effect downstream on pose estimates where uncertainty is higher. For example with our rosette-shaped trajectories, a small pose error near the central landmark could drift to a much larger error as the robot moves away from the center and errors accumulate. To propagate these local refinements globally, we incorporate the refined poses as 3DGS-derived factors in our factor graph. For each refined pose $i \in \Iext$, we add a cost term:

\vspace{-0.4cm}
\begin{equation}
J_{\mathrm{ext}}(\mathcal{X})
=
\sum_{i \in \Iext}
\wnorm{
\Logmap\!\Big( (\Xext{i})^{-1}\,\X{i} \Big)
}{
\Sigext{i}
}
\label{eq:external_pose_priors}
\end{equation}

\noindent where $\Xext{i}$ is the pose refined for a pose variable $\X{i}$ and $\Sigext{i}$ encodes the predefined uncertainty of the refined poses. The sets of final poses $\mathcal{X}^{\star\star}$ and landmark estimates $\mathcal{L}^{\star\star}$ are obtained by jointly optimizing the original factor graph objective and the additional external pose constraints:
\begin{equation}
\mathcal{X}^{\star\star}, \mathcal{L}^{\star\star}
=
\arg\min_{\mathcal{X},\,\mathcal{L}}
\left(
J_{\mathrm{base}}(\mathcal{X},\mathcal{L})
+
J_{\mathrm{3DGS}}(\mathcal{X})
\right)
\label{eq:refined_map}
\end{equation}

The addition of these 3DGS camera factors can cause a large shift in the global trajectory, as shown in \cref{fig:pose_refinement} from light gray to dark purple, particularly in uncertain regions far from the landmark. Thus, before expanding the frontier of the 3DGS model and incorporating a new set of image observations, we perform this global trajectory optimization. Specifically, we only perform global trajectory optimization if the average uncertainty of camera poses at the frontier region has not increased significantly, more than doubled, from the central, lowest uncertainty seed region. In other words, when the factor graph is most certain about robot states, but these state estimates do not align with the 3DGS scene, we must incorporate our additional 3DGS camera factors to correct these biases.

We optimize our 3DGS map, $\mathcal{G}$, with a modified uncertainty-aware version of the original 3DGS reconstruction loss \cite{kerbl_3d_2023} and an additional edge-aware total variation loss \cite{godard2017unsupervised} on the rendered depth, $\loss_{\text{Z}_\text{TV}}$, promoting smoothness while allowing for variation along areas of high gradient in the observed camera image (e.g., edges). We incorporate a DINOv2 \cite{oquab_dinov2_2024} based uncertainty modeling component similar to past works \cite{zheng_wildgs-slam_2025, kulhanek_wildgaussians_2024, ren_nerf_2024} by projecting 3D-aware features \cite{yue_improving_2024} through a shallow MLP, $\mathcal{P}$, to an uncertainty map, $\beta$, with modulates the 3DGS L1 reconstruction loss:

\vspace{-0.5cm}
\begin{align}
    \label{eq:loss-recon}
    \loss_{\mathcal{G}} = (1 - \lambda_{\text{1}})\Big|\Big| \frac{I - \hat{I}}{\beta_\text{detach}} \Big|\Big|_1 + \lambda_{\text{1}} \loss_{\text{SSIM}} + \lambda_2 \loss_{\text{Z}_\text{TV}}
\end{align}

The uncertainty MLP, ${\mathcal{P}}$, is optimized similar to \cite{zheng_wildgs-slam_2025}, with the modified SSIM loss and two regularization terms from \cite{ren_nerf_2024}. One regularization term, $\loss_{\text{unc-var}}$, minimizes the variance of uncertainty for similar features, while the other, $\loss_{\text{unc-log}} = \log\beta$, prevents $\beta$ from a degenerate solution of high uncertainty everywhere. ${\mathcal{P}}$ is jointly optimized with the 3DGS map, ${\mathcal{G}}$, with the $\beta$ detached in the reconstruction loss and renders from $\mathcal{G}$ detached in the uncertainty loss.

\vspace{-0.4cm}
\begin{equation}
    \label{eq:loss-uncertainty}
    \loss_{\mathcal{P}} = \frac{\loss^{'}_{\text{SSIM}}}{\beta^2} + \lambda_3 \loss_{\text{unc-var}} + \lambda_4 \loss_{\text{unc-log}}
\end{equation}

\subsection{Using monodepth estimators to generate pseudo depth}
While 3DGS normally relies on initialization from a point cloud, typically a byproduct of SfM, or depends on expensive sampling through methods like MCMC \cite{kheradmand_3d_nodate}, \Method avoids using structure-from-motion entirely. We use a DepthAnythingV2 \cite{yang_depth_2024} model fine-tuned to output metric depth for our underwater benthic imagery. We generate a pseudo ground truth dataset to fine-tune this model by fitting dense depth from an off-the-shelf relative depth pre-trained DepthAnythingV2 model with the sparse point cloud obtained from Metashape, similar to previous works \cite{ turkulainen2024dnsplatter}. In the absence of stereo vision or a depth sensor on our AUV, we use the metric depth model's outputs to initialize new Gaussians in our 3DGS scene.

\section{Experiments}
\label{sec:exp}

\subsection{Dataset collection}
We evaluate \Method on coral reef benthic survey data collected at two sites in the US Virgin Islands, Tektite and Yawzi. Both sites are known to be biologically active and contain a patchy mix of sand, seagrass, and hard and soft corals \cite{aoki_replayed_2024}. An AprilTag \cite{olson2011tags} is placed at the center of the reef, and CUREE \cite{girdhar_curee_2023}, a flexible low-cost underwater vehicle platform, collects sensor data. CUREE has a sensor suite including a 3 Hz 4K downward-facing monocular camera with a FOV of 120\textdegree$\times$58\textdegree in a dome port, a Waterlinked DVL A50 (4-15 Hz), an ICM-20602 IMU from a BlueRobotics Navigator (100 Hz), and a BlueRobotics Bar30 pressure sensor (22 Hz). Fixed rosette trajectories were generated for CUREE to follow with its navigation system while aiming to maintain an altitude of 2 m above the seafloor. Half of a rosette trajectory, spanning 347.3 m over 27.0 min, is executed at Tektite, while a full rosette trajectory, spanning 695.5 m over 43.2 min, is executed at Yawzi.

The RAW linear imagery is white-balanced and converted to sRGB. With the consistent survey altitude and downward-facing imagery (e.g., minimal, consistent water column and distance-dependent effects such as attenuation and backscatter), we find this results in visually consistent data. However, unlike typical room-scale datasets, where there is a high degree of overlap between frames, our top-down captured benthic imagery shows significantly less overlap. We rectify the images using the vehicle's default calibration to mitigate distortion effects. We further utilize only the center crop of the image, maintaining the same aspect ratio, for optimizing our 3D scene representations as this region is less distorted than the periphery. This cropping further minimizes the visual overlap of frames as compared to typical visual SLAM datasets. Sample images are shown in \cref{fig:intro}.

\subsection{Evaluation metrics}
We evaluate our framework through two different criteria, visual fidelity of the observed images in the scene reconstruction and absolute trajectory error (ATE) of the robot trajectory. Visual fidelity is evaluated with peak signal-to-noise ratio (PSNR), structural similarity index (SSIM), and perceptual distance (LPIPS). The ground truth reference of the robot trajectory is approximated using Metashape to estimate the pose of each image collected, and we use evo \cite{grupp2017evo} to align the reference before computing the ATE RMSE in meters. We also report the trajectory length, as not all methods successfully track the full image sequence, providing context for interpreting the ATE RMSE metric.

\subsection{Comparative baselines}

We evaluate our framework against other visual SLAM methods: ORB-SLAM3 \cite{campos_orb-slam3_2021}, DROID-SLAM \cite{teed_droid-slam_2022}, MASt3R-SLAM \cite{murai_mast3r-slam_2024}, VGGT-SLAM \cite{maggio2025vggt-slam}, MonoGS \cite{matsuki_gaussian_2024}, WildGS-SLAM \cite{zheng_wildgs-slam_2025}. With these methods, we focus on the ATE RMSE, as not all methods provide a dense reconstruction from which reconstruction quality can be quantitatively evaluated. We also evaluate against a baseline multimodal landmark SLAM implementation with GTSAM \cite{gtsam}, suggesting an upper bound on ATE RMSE using solely multimodal sensor data without integrating dense visual information.

To measure reconstruction quality, we evaluate our framework with various permutations of input data, specifically the camera poses and initialization points, and the 3DGS optimization procedure, either in batch where all images and poses are available initially or incrementally, following the radially outward optimization from region of lowest to highest pose uncertainty as described in \cref{subsec:incremental}. Some input permutations, those utilizing poses or the pointcloud from Metashape, are unrealistic to obtain in an ``in the wild" field robotics context and are thus considered to use oracle information. We also ablate our framework with and without the global optimization step, where 3DGS camera factors are added into the pose graph to re-optimize the trajectory. We annotate each of the methods in \cref{tab:recon-metrics} as such. For example, Metashape+sfm+batch means poses from Metashape, initialization points from SfM, and 3DGS optimization in batch as typically done. On the other hand, GTSAM+md+inc means camera poses from our factor graph, initialization points from a depth estimator, and 3DGS optimization incrementally. As an ablation, +reopt indicates if global reoptimization of the factor graph occurs during incremental 3DGS optimization. We also compare reconstruction quality with the WildGS-SLAM \cite{zheng_wildgs-slam_2025} output. All experiments are run with a Ryzen 9 7900X CPU and RTX 6000 Ada GPU.
\section{Results}
\label{sec:result}

\begin{table}[tbp]
    \centering
    \begin{tabular}{l|cc|cc}
        \toprule
        \multicolumn{1}{l|}{} & \multicolumn{2}{c|}{Tektite} & \multicolumn{2}{c}{Yawzi} \\
        \scriptsize Method & \scriptsize Length (m) & \scriptsize RMSE (m) & \scriptsize Length (m) & \scriptsize RMSE (m) \\
        \midrule
        \cg Metashape (ref)            & \cg 347.305 & \cg 0.000 & \cg 695.504 & \cg 0.000 \\
        GTSAM                          & \fst 347.198 & \rd 0.328 & \fst 701.151 & \rd 0.283 \\
        ORB-SLAM3                      & 17.714 & \nd 0.202 & 6.931 & \fst 0.222\\
        DROID-SLAM                     & 18.343 & 4.305 & 142.180 & 4.532 \\
        VGGT-SLAM                      & 11.938 & 5.029 & 83.476 & 6.085 \\
        MASt3R-SLAM                    & 52.540 & 5.496 & 48.608 & 1.398 \\
        MonoGS                         & - & f & - & f \\
        WildGS-SLAM                    & \rd 139.286 & 4.408 & \nd 681.151 & 4.488 \\
        \textbf{\Method}  & \nd 361.458 & \fst 0.135 & \rd 721.253 & \nd 0.229 \\
        \bottomrule
    \end{tabular}
        
    \caption{\textbf{Trajectory estimation metrics.} ATE RMSE is calculated in meters. Some methods are unable to track all frames, so we report RMSE over those frames that are tracked as well as total path length. Lower RMSE is better but the path length should also be closer to the reference. Best results using non-oracle data are highlighted as \colorbox{colorFst}{first}, \colorbox{colorSnd}{second}, and \colorbox{colorTrd}{third}. f denotes complete tracking failure. For RMSE, lower is better $\downarrow.$}    
    \label{tab:rmse-metrics}
\end{table}

\begin{figure}[t]
        \centering
        \includegraphics[width=\linewidth]{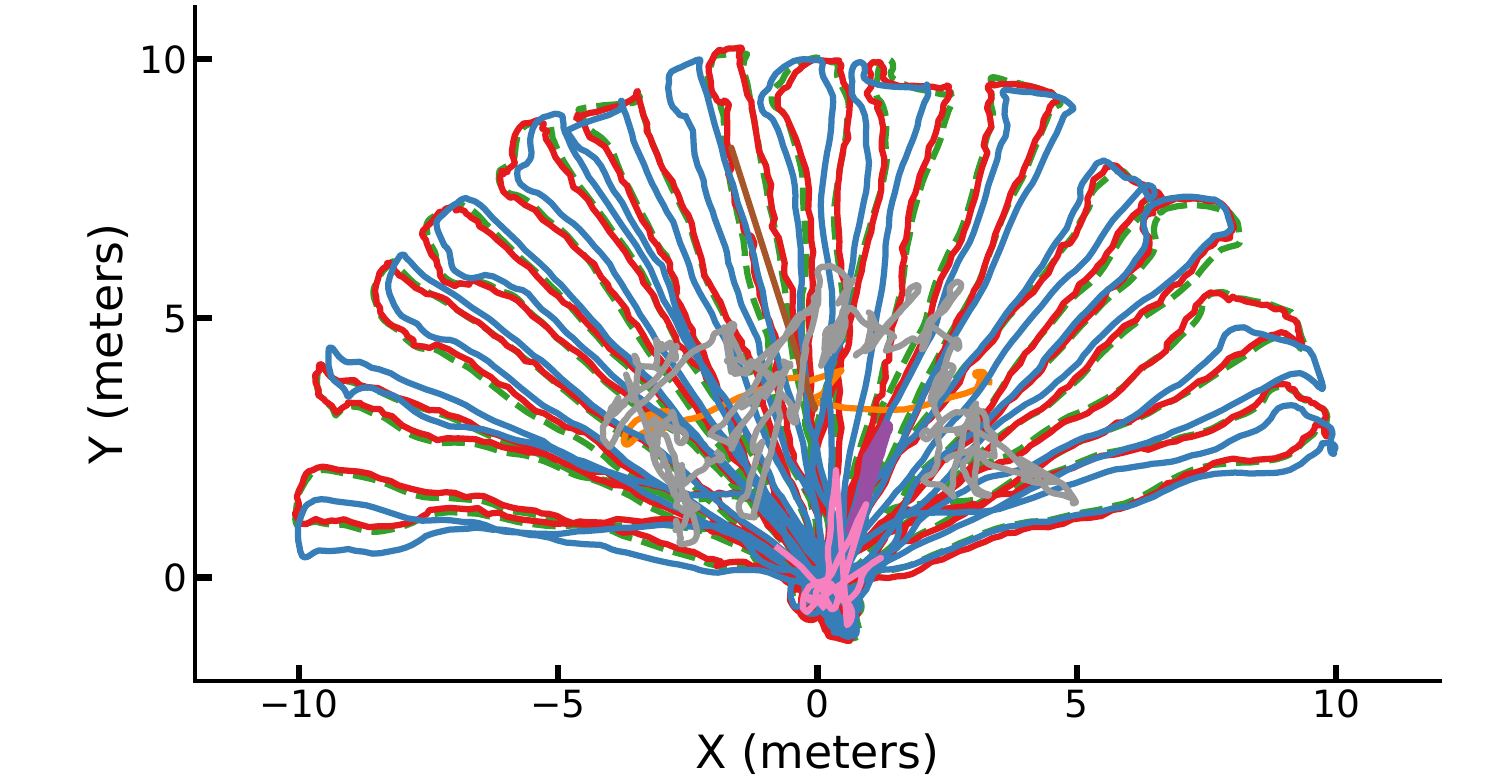}
        \includegraphics[width=\linewidth]{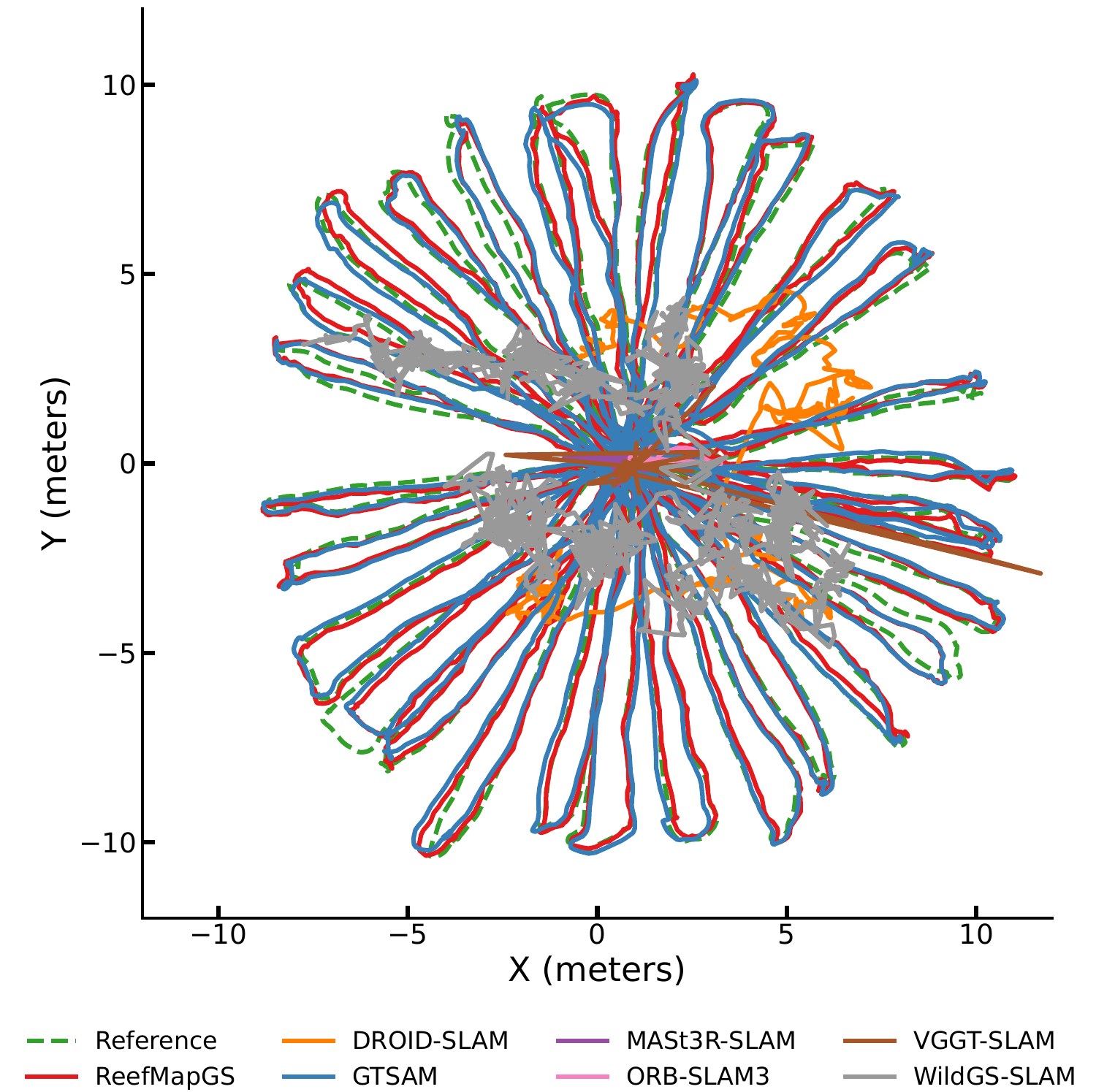}
        \caption{\textbf{Qualitative trajectory evaluation} Top-down view of robot trajectories estimated by \Method and other baselines methods are plotted against the reference trajectory estimated by Metashape, dashed green, at both reef sites, Tektite (above) and Yawzi (below).}
        \label{fig:2d-trajs}
\end{figure}

\subsection{Pose estimation accuracy}

Average trajectory error RMSE as well as the estimated path length, in meters, are shown for all methods quantitatively in \cref{tab:rmse-metrics} and qualitatively in \cref{fig:2d-trajs}. Ideally, we want a method with low RMSE and closer to reference path length. For ATE RMSE, we observe that \Method has the lowest RMSE on the Tektite scene (0.135 m) and the second-lowest on the Yawzi scene (0.229 m), while also closely estimating the trajectory length. While ORB-SLAM3 improves upon the RMSE by less than a centimeter on Yawzi, it struggles to run the entire trajectory, estimating only a few meters of it. Importantly, \Method improves noticeably upon the ATE RMSE estimated by GTSAM by 58.8\% on Tektite and 19.0\% on Yawzi. In other words, incorporating a dense 3DGS-based map representation to provide additional signal in pose graph-based multimodal SLAM, as \Method does, helps more accurately estimate the whole robot trajectory. 

Learning-based methods all encounter significant issues in estimating the robot's trajectory. Likewise, 3DGS-based systems struggle or are unable to initialize. We note that the RMSEs for all these methods are an order of magnitude larger, in the 4-5 m range, which is significant relative to the 10 m rosette radius. They have difficulty tracking the robot, tracking for a few frames or sections of smoother, straight motion before losing tracking, and perhaps recovering for a later frame. They especially have difficulty with the sharp turn at the end of each rosette petal, suggesting an area of improvement in survey trajectory design for smoother, rounded petals. We hypothesize that learning-based methods struggle with the nature of our data: its top-down viewpoint, narrow field of view, and unique semantic content all make it out-of-distribution from typical training data. ORB-SLAM3 tracks a few meters of the trajectory near the center, with an ATE RMSE competitive with \Method, before losing tracking until the robot later returns, highlighting that classical pipelines still have value in the era of learning-based methods. We also note that GTSAM and \Method provide close to metric-scale trajectories as they leverage inertial, velocity, and depth measurements whereas other methods rely purely on visual features and are off by a scale factor. 

\begin{table*}[ht]
    \centering
    \begin{tabular}{l|l|cccc|cccc}
        \toprule
        \multicolumn{1}{l|}{} &\multicolumn{1}{l|}{} & \multicolumn{4}{c|}{Tektite} & \multicolumn{4}{c}{Yawzi} \\
        \multicolumn{1}{l|}{} & Method & PSNR $\uparrow$ & SSIM $\uparrow$ & LPIPS $\downarrow$ & RMSE (m) $\downarrow$ & PSNR $\uparrow$ & SSIM $\uparrow$ & LPIPS $\downarrow$ & RMSE (m) $\downarrow$ \\
        \midrule
        \cg
        \parbox[t]{1mm}{\multirow{4}{*}{\rotatebox[origin=c]{90}{Oracle}}} 
            & \cg Metashape+sfm+batch                 & \cg 20.447 & \cg 0.499 & \cg 0.886 & \cg \textbf{0.000} & \cg 19.977 & \cg 0.492 & \cg 0.918 & \cg \textbf{0.000} \\
            & \cg Metashape+sfm+inc                   & \cg \textbf{22.395} & \cg \textbf{0.577} & \cg \textbf{0.673} & \cg 0.070 & \cg \textbf{22.522} & \cg \textbf{0.594} & \cg \textbf{0.680} & \cg 0.064 \\
            & \cg Metashape+md+batch                  & \cg 20.273 & \cg 0.501 & \cg 0.880 & \cg \textbf{0.000} & \cg 20.060 & \cg 0.496 & \cg 0.917 & \cg \textbf{0.000} \\
            & \cg Metashape+md+inc                    & \cg 21.975 & \cg 0.564 & \cg 0.702 & \cg 0.066 & \cg 22.158 & \cg 0.583 & \cg 0.701 & \cg 0.059 \\
        \midrule
        \parbox[t]{1mm}{\multirow{4}{*}{\rotatebox[origin=c]{90}{In the wild}}} 
            & GTSAM+md+batch                      & 17.846 & 0.461 & 0.945 & \nd 0.328 & 17.683 & 0.467 & 0.958 & \nd 0.283 \\
            & GTSAM+md+inc                        & \nd 21.051 & \nd 0.542 & \nd 0.767 & 0.336 & \nd 20.746 & \nd 0.541 & \nd 0.799 & 0.331 \\
            & WildGS-SLAM                         & 10.253 & 0.242 & 0.980 & 4.407 & 9.779 & 0.272 & 0.993 & 4.488 \\
            & \textbf{ReefMapGS (GTSAM+md+inc+reopt)}  & \fst 22.219 & \fst 0.567 & \fst 0.721 & \fst 0.135 & \fst 21.226 & \fst 0.551 & \fst 0.777 & \fst 0.229 \\
            \bottomrule
    \end{tabular}
        
    \caption{\textbf{Reconstruction and tracking metrics across different methods.} We ablate across different configurations of input data and 3DGS optimization and with WildGS-SLAM. Initial poses can come from \texttt{GTSAM} or \texttt{Metashape}, Gaussians are initialized either from the SfM point cloud, \texttt{sfm}, or depth estimator outputs, \texttt{md}, and the 3DGS can be optimized either in batch, \texttt{batch}, or incrementally following our method, \texttt{inc}. Input data configurations that use oracle data from SfM are {\cg grayed}. Best results using non-oracle data are highlighted as \colorbox{colorFst}{first} and \colorbox{colorSnd}{second}, while best results among all are \textbf{bolded}.}
    \label{tab:recon-metrics}
\end{table*}

\begin{table}[htbp]
    \centering
    \begin{tabular}{l|cccc|cccc}
        \toprule
        \multicolumn{1}{l|}{} & \multicolumn{2}{c}{\scriptsize Runtime (hh:mm)} \\
        Method & Tektite & Yawzi \\
        \midrule
        COLMAP                              & 10:40 & 17:14 \\
        Metashape                           & 8:04 & 6:57 \\
        ORB-SLAM3                           & 0:01 & 0:02 \\
        DROID-SLAM                          & 0:03 & 0:05 \\
        VGGT-SLAM                           & 0:08 & 0:06 \\
        MASt3R-SLAM                         & 0:14 & 0:14 \\
        WildGS-SLAM                         & 8:36 & 8:41 \\
        \textbf{ReefMapGS}                       & 2:30 & 3:22 \\
        \bottomrule
    \end{tabular}
        
    \caption{\textbf{Processing time.} \Method cuts down processing time significantly compared to open source (COLMAP) and proprietary (Metashape) SfM software, requiring roughly 3 hours for both scenes. While learning-based methods are fast none yield coherent results on our scenes.}
    \label{tab:runtime-metrics}
\end{table}

\subsection{3D reconstruction quality}
\cref{tab:recon-metrics} shows the reconstruction quality of \Method and baselines across both the Tektite and Yawzi scenes. Across both scenes, \Method achieves higher reconstruction quality than when using the oracle data with batch optimization, Metashape+sfm+batch. When incorporating our incremental optimization with local pose refinement as new frontiers are explored (+inc instead of +batch), we observe that, across all input data scenarios, even when using only SfM data, reconstruction quality increases significantly. This highlights the effectiveness of our incremental 3DGS optimization, ensuring scene coherence by optimizing it from the most certain region and aligning new observations with the reconstruction thus far.

We also see that initializing Gaussians with depth estimator data yields a slight, but not catastrophic, decrease in quality, as opposed to initializing Gaussians with the SfM point cloud (+md instead of +sfm). This highlights the role that depth images can use in 3DGS-based pipelines for visual SLAM and shows that monocular depth estimators can be a powerful replacement when stereo vision is unavailable, either from hardware limitations or when visual data quality prevents application of stereo matching and depth from disparity algorithms (e.g., low texture sandy regions).  

Finally, the results show that global re-optimization of the factor graph using 3DGS camera pose factors further increases reconstruction quality. This highlights how incorporating 3DGS camera pose factors, resulting from local camera refinement while incrementally optimizing the model, can increase model quality. Intuitively, if two petals of the rosette have overlapping visual content, but the trajectories have drifted apart due to an error in pose estimation closer to the central landmark, then local pose refinement may not perform this visual loop closure. With global pose refinement incorporating small corrections that accumulate to these large drifts, redundant visual structures from different trajectory petals can be reconciled, increasing reconstruction quality.  

\subsection{Running time}
\cref{tab:runtime-metrics} shows the different runtimes for various methods. Note that for COLMAP and Metashape, we only report the time required for camera pose estimation, not for 3D reconstruction. We observe that SfM pipelines require substantial time, with COLMAP requiring approximately 11 and 17 hours for the Tektite and Yawzi scenes, respectively. While commercial software reduces the runtime to 8 and 7 hours, respectively, \Method requires even less time, with approximately 3 hours for both scenes. The local pose refinement process is the slowest component of \Method, requiring multiple iterations of gradient descent per observation. Notably, \Method is 2-3x faster than the other 3DGS-based system. However, we see that learning-based and classical methods are able to process the entire scene even faster, requiring mere minutes. While none of the learning-based methods yielded coherent outputs on our datasets, this highlights an exciting and promising future direction.

\section{Conclusions}
\label{sec:conclusions}
We present \Method, a framework that tightly integrates multimodal pose-graph SLAM with incremental 3DGS for large-scale underwater scene reconstruction. Starting from pose estimates obtained by fusing IMU, DVL, depth, and landmark measurements, we incrementally build a dense 3DGS representation and constrain pose-graph SLAM. As we integrate new information, we locally refine camera poses estimates through differentiable rendering with our 3DGS scene and propagate those refinements globally by adding factors back into the pose-graph and optimizing. This approach enables pose estimation and scene reconstruction to mutually improve. Our approach reconstructs large-scale underwater scenes without relying on and faster than computationally intensive SfM, making it more useful in remote, field deployment contexts. We show high-quality 3D reconstruction performance, competitive with or better than when using oracle SfM data as inputs or against another 3DGS-based SLAM method. We also show low ATE, both against a multimodal SLAM baseline that does not integrate information from a dense scene representation model as well as against other visual SLAM methods that struggle to yield coherent output on our challenging datasets. Future work includes extending the framework to surveys without fixed landmarks, exploring strategies for global re-optimization, and improving \Method to enable deployment on a robot. Overall, \Method provides a practical path toward reliable, large-scale underwater mapping by closing the loop between multimodal SLAM and 3DGS.

{
\bibliographystyle{IEEEtran}
\bibliography{ref, ref-clean}

@inproceedings{bian_unsupervisedscale_2019,
	title = {Unsupervised {Scale}-consistent {Depth} and {Ego}-motion {Learning} from {Monocular} {Video}},
	volume = {32},
	urldate = {2026-01-15},
	booktitle = {Advances in {Neural} {Information} {Processing} {Systems}},
	author = {Bian, Jiawang and Li, Zhichao and Wang, Naiyan and Zhan, Huangying and Shen, Chunhua and Cheng, Ming-Ming and Reid, Ian},
	year = {2019},
}

@INPROCEEDINGS{girdhar_curee_2023,
  author={Girdhar, Yogesh and McGuire, Nathan and Cai, Levi and Jamieson, Stewart and McCammon, Seth and Claus, Brian and Soucie, John E. San and Todd, Jessica E. and Mooney, T. Aran},
  booktitle={2023 IEEE International Conference on Robotics and Automation}, 
  title={CUREE: A Curious Underwater Robot for Ecosystem Exploration}, 
  year={2023},
  volume={},
  number={},
  pages={11411-11417},
  doi={10.1109/ICRA48891.2023.10161282}
}

@inproceedings{song_turtlmap_2024,
  author={Song, Jingyu and Bagoren, Onur and Andigani, Razan and Sethuraman, Advaith and Skinner, Katherine A.},
  booktitle={2024 IEEE/RSJ International Conference on Intelligent Robots and Systems (IROS)}, 
  title={TURTLMap: Real-time Localization and Dense Mapping of Low-texture Underwater Environments with a Low-cost Unmanned Underwater Vehicle}, 
  year={2024},
  volume={},
  number={},
  pages={1191-1198},
  doi={10.1109/IROS58592.2024.10801692}
}

@inproceedings{williams_simultaneous_2004,
  author={Williams, S. and Mahon, I.},
  booktitle={IEEE International Conference on Robotics and Automation, 2004. Proceedings. ICRA '04. 2004}, 
  title={Simultaneous localisation and mapping on the Great Barrier Reef}, 
  year={2004},
  volume={2},
  number={},
  pages={1771-1776 Vol.2},
  doi={10.1109/ROBOT.2004.1308080}}

@INPROCEEDINGS{keetha_splatam_2024,
  author={Keetha, Nikhil and Karhade, Jay and Jatavallabhula, Krishna Murthy and Yang, Gengshan and Scherer, Sebastian and Ramanan, Deva and Luiten, Jonathon},
  booktitle={2024 IEEE/CVF Conference on Computer Vision and Pattern Recognition (CVPR)}, 
  title={SplaTAM: Splat, Track and Map 3D Gaussians for Dense RGB-D SLAM}, 
  year={2024},
  volume={},
  number={},
  pages={21357-21366},
  doi={10.1109/CVPR52733.2024.02018}
}

@INPROCEEDINGS{zheng_wildgs-slam_2025,
  author={Zheng, Jianhao and Zhu, Zihan and Bieri, Valentin and Pollefeys, Marc and Peng, Songyou and Armeni, Iro},
  booktitle={2025 IEEE/CVF Conference on Computer Vision and Pattern Recognition (CVPR)}, 
  title={WildGS-SLAM: Monocular Gaussian Splatting SLAM in Dynamic Environments}, 
  year={2025},
  volume={},
  number={},
  pages={11461-11471},
  doi={10.1109/CVPR52734.2025.01070}}

@INPROCEEDINGS{wang_dust3r_2024,
  author={Wang, Shuzhe and Leroy, Vincent and Cabon, Yohann and Chidlovskii, Boris and Revaud, Jerome},
  booktitle={2024 IEEE/CVF Conference on Computer Vision and Pattern Recognition}, 
  title={DUSt3R: Geometric 3D Vision Made Easy}, 
  year={2024},
  volume={},
  number={},
  pages={20697-20709},
  doi={10.1109/CVPR52733.2024.01956}}

@INPROCEEDINGS{wang_vggt_2025,
  author={Wang, Jianyuan and Chen, Minghao and Karaev, Nikita and Vedaldi, Andrea and Rupprecht, Christian and Novotny, David},
  booktitle={2025 IEEE/CVF Conference on Computer Vision and Pattern Recognition (CVPR)}, 
  title={VGGT: Visual Geometry Grounded Transformer}, 
  year={2025},
  volume={},
  number={},
  pages={5294-5306},
  doi={10.1109/CVPR52734.2025.00499}}

@article{teed_droid-slam_2022,
  title={{DROID-SLAM: Deep Visual SLAM for Monocular, Stereo, and RGB-D Cameras}},
  author={Teed, Zachary and Deng, Jia},
  journal={Advances in neural information processing systems},
  year={2021}
}

@inproceedings{murai_mast3r-slam_2024,
  title={{MASt3R-SLAM}: Real-Time Dense {SLAM} with {3D} Reconstruction Priors},
  author={Murai, Riku and Dexheimer, Eric and Davison, Andrew J.},
  booktitle={Proceedings of the IEEE/CVF Conference on Computer Vision and Pattern Recognition},
  year={2025},
}

@article{aoki_replayed_2024,
  title={Replayed reef sounds induce settlement of Favia fragum coral larvae in aquaria and field environments},
  author={Aoki, Nad{\`e}ge and Weiss, Benjamin and J{\'e}z{\'e}quel, Youenn and Apprill, Amy and Mooney, T Aran},
  journal={JASA Express Letters},
  volume={4},
  number={10},
  year={2024},
  publisher={AIP Publishing}
}

@article{kerbl_3d_2023,
  title={3d gaussian splatting for real-time radiance field rendering.},
  author={Kerbl, Bernhard and Kopanas, Georgios and Leimk{\"u}hler, Thomas and Drettakis, George and others},
  journal={ACM Trans. Graph.},
  volume={42},
  number={4},
  pages={139--1},
  year={2023}
}

@inproceedings{duisterhof_mast3r-sfm_2024,
  title={Mast3r-sfm: a fully-integrated solution for unconstrained structure-from-motion},
  author={Duisterhof, Bardienus Pieter and Zust, Lojze and Weinzaepfel, Philippe and Leroy, Vincent and Cabon, Yohann and Revaud, Jerome},
  booktitle={2025 International Conference on 3D Vision (3DV)},
  pages={1--10},
  year={2025},
  organization={IEEE}
}

@article{sauder_scalable_2024,
  title={Scalable semantic 3D mapping of coral reefs with deep learning},
  author={Sauder, Jonathan and Banc-Prandi, Guilhem and Meibom, Anders and Tuia, Devis},
  journal={Methods in Ecology and Evolution},
  volume={15},
  number={5},
  pages={916--934},
  year={2024},
  publisher={Wiley Online Library}
}

@article{yang_depth_2024,
 author = {Yang, Lihe and Kang, Bingyi and Huang, Zilong and Zhao, Zhen and Xu, Xiaogang and Feng, Jiashi and Zhao, Hengshuang},
 title = {Depth Anything V2},
 journal={Advances in Neural Information Processing Systems},
 volume={37},
 year={2024}
}

@article{maggio2025vggt-slam,
  title={VGGT-SLAM: Dense RGB SLAM Optimized on the SL (4) Manifold},
  author={Maggio, Dominic and Lim, Hyungtae and Carlone, Luca},
  journal={Advances in Neural Information Processing Systems},
  volume={39},
  year={2025}
}

@article{campos_orb-slam3_2021,
  title={Orb-slam3: An accurate open-source library for visual, visual--inertial, and multimap slam},
  author={Campos, Carlos and Elvira, Richard and Rodr{\'\i}guez, Juan J G{\'o}mez and Montiel, Jos{\'e} MM and Tard{\'o}s, Juan D},
  journal={IEEE transactions on robotics},
  volume={37},
  number={6},
  pages={1874--1890},
  year={2021},
  publisher={IEEE}
}

@misc{oquab_dinov2_2024,
  title={DINOv2: Learning Robust Visual Features without Supervision},
  author={Oquab, Maxime and Darcet, Timothée and Moutakanni, Theo and Vo, Huy V. and Szafraniec, Marc and Khalidov, Vasil and Fernandez, Pierre and Haziza, Daniel and Massa, Francisco and El-Nouby, Alaaeldin and Howes, Russell and Huang, Po-Yao and Xu, Hu and Sharma, Vasu and Li, Shang-Wen and Galuba, Wojciech and Rabbat, Mike and Assran, Mido and Ballas, Nicolas and Synnaeve, Gabriel and Misra, Ishan and Jegou, Herve and Mairal, Julien and Labatut, Patrick and Joulin, Armand and Bojanowski, Piotr},
  journal={arXiv:2304.07193},
  year={2023}
}

@article{kulhanek_wildgaussians_2024,
    title={{W}ild{G}aussians: {3D} Gaussian Splatting in the Wild},
    author={Kulhanek, Jonas and Peng, Songyou and Kukelova, Zuzana and Pollefeys, Marc and Sattler, Torsten},
    journal={Advances in Neural Information Processing Systems},
    volume={38},
    year={2024}
}

@InProceedings{ren_nerf_2024,
  author={Ren, Weining and Zhu, Zihan and Sun, Boyang and Chen, Jiaqi and Pollefeys, Marc and Peng, Songyou},
  booktitle={2024 IEEE/CVF Conference on Computer Vision and Pattern Recognition (CVPR)}, 
  title={NeRF On-the-go: Exploiting Uncertainty for Distractor-free NeRFs in the Wild}, 
  year={2024},
  volume={},
  number={},
  pages={8931-8940},
  doi={10.1109/CVPR52733.2024.00853}}

@inproceedings{yue_improving_2024,
  title     = {{Improving 2D Feature Representations by 3D-Aware Fine-Tuning}},
  author    = {Yue, Yuanwen and Das, Anurag and Engelmann, Francis and Tang, Siyu and Lenssen, Jan Eric},
  booktitle = {European Conference on Computer Vision (ECCV)},
  year      = {2024}
}

@article{kheradmand_3d_nodate,
	title = {{3D} {Gaussian} {Splatting} as {Markov} {Chain} {Monte} {Carlo}},
	author = {Kheradmand, Shakiba and Rebain, Daniel and Sharma, Gopal and Sun, Weiwei and Tseng, Yang-Che and Isack, Hossam and Kar, Abhishek and Tagliasacchi, Andrea and Yi, Kwang Moo},
    journal={Advances in Neural Information Processing Systems},
    volume={38},
    year={2024}
}

@article{wang2024spann3r,
    title={3D Reconstruction with Spatial Memory},
    author={Wang, Hengyi and Agapito, Lourdes},
    journal={arXiv preprint arXiv:2408.16061},
    year={2024}
}

@INPROCEEDINGS{wang2025cut3r,
  author={Wang, Qianqian and Zhang, Yifei and Holynski, Aleksander and Efros, Alexei A. and Kanazawa, Angjoo},
  booktitle={2025 IEEE/CVF Conference on Computer Vision and Pattern Recognition (CVPR)}, 
  title={Continuous 3D Perception Model with Persistent State}, 
  year={2025},
  volume={},
  number={},
  pages={10510-10522},
  doi={10.1109/CVPR52734.2025.00983}}

@inproceedings{matsuki_gaussian_2024,
  title={{G}aussian {S}platting {SLAM}},
  author={Hidenobu Matsuki and Riku Murai and Paul H. J. Kelly and Andrew J. Davison},
  booktitle={Proceedings of the IEEE/CVF Conference on Computer Vision and Pattern Recognition},
  year={2024}
}

@misc{noauthor_agisoft_nodate,
	title = {Agisoft {Metashape}: {Professional} {Edition}},
	url = {https://www.agisoft.com/features/professional-edition/},
	urldate = {2023-04-28},
}

@InProceedings{turkulainen2024dnsplatter,
  author={Turkulainen, Matias and Ren, Xuqian and Melekhov, Iaroslav and Seiskari, Otto and Rahtu, Esa and Kannala, Juho},
  booktitle={2025 IEEE/CVF Winter Conference on Applications of Computer Vision (WACV)}, 
  title={DN-Splatter: Depth and Normal Priors for Gaussian Splatting and Meshing}, 
  year={2025},
  volume={},
  number={},
  pages={2421-2431},
  doi={10.1109/WACV61041.2025.00241}}

@article{valenti2015keeping,
  title={Keeping a good attitude: A quaternion-based orientation filter for IMUs and MARGs},
  author={Valenti, Roberto G and Dryanovski, Ivan and Xiao, Jizhong},
  journal={Sensors},
  volume={15},
  number={8},
  pages={19302--19330},
  year={2015},
  publisher={Multidisciplinary Digital Publishing Institute}
}

@inproceedings{moore2016generalized,
  title={A generalized extended kalman filter implementation for the robot operating system},
  author={Moore, Thomas and Stouch, Daniel},
  booktitle={Intelligent Autonomous Systems 13: Proceedings of the 13th International Conference IAS-13},
  pages={335--348},
  year={2016},
  organization={Springer}
}

@inproceedings{wang2023real,
  title={Real-time dense 3d mapping of underwater environments},
  author={Wang, Weihan and Joshi, Bharat and Burgdorfer, Nathaniel and Batsosc, Konstantinos and Lid, Alberto Quattrini and Mordohaia, Philippos and Rekleitisb, Ioannis},
  booktitle={2023 IEEE International Conference on Robotics and Automation (ICRA)},
  pages={5184--5191},
  year={2023},
  organization={IEEE}
}

@article{xu2025aqua,
  title={AQUA-SLAM: Tightly-Coupled Underwater Acoustic-Visual-Inertial SLAM with Sensor Calibration},
  author={Xu, Shida and Zhang, Kaicheng and Wang, Sen},
  journal={IEEE Transactions on Robotics},
  year={2025},
  publisher={IEEE}
}

@inproceedings{joshi2019experimental,
  title={Experimental comparison of open source visual-inertial-based state estimation algorithms in the underwater domain},
  author={Joshi, Bharat and Rahman, Sharmin and Kalaitzakis, Michail and Cain, Brennan and Johnson, James and Xanthidis, Marios and Karapetyan, Nare and Hernandez, Alan and Li, Alberto Quattrini and Vitzilaios, Nikolaos and others},
  booktitle={2019 IEEE/RSJ International Conference on Intelligent Robots and Systems (IROS)},
  pages={7227--7233},
  year={2019},
  organization={IEEE}
}

@inproceedings{olson2011tags,
    TITLE      = {{AprilTag}: A robust and flexible visual fiducial system},
    AUTHOR     = {Edwin Olson},
    BOOKTITLE  = {Proceedings of the {IEEE} International Conference on Robotics and
                 Automation ({ICRA})},
    YEAR       = {2011},
    MONTH      = {May},
    PAGES      = {3400-3407},
    PUBLISHER  = {IEEE},
}

@article{ye2025gsplat,
  title={gsplat: An open-source library for Gaussian splatting},
  author={Ye, Vickie and Li, Ruilong and Kerr, Justin and Turkulainen, Matias and Yi, Brent and Pan, Zhuoyang and Seiskari, Otto and Ye, Jianbo and Hu, Jeffrey and Tancik, Matthew and others},
  journal={Journal of Machine Learning Research},
  volume={26},
  number={34},
  pages={1--17},
  year={2025}
}

@String(CVPR= {IEEE Conf. Comput. Vis. Pattern Recog.})

@String(ECCV= {Eur. Conf. Comput. Vis.})

@String(CVPR  = {CVPR})

@String(ECCV  = {ECCV})

@inproceedings{schoenberger2016sfm,
    author={Sch\"{o}nberger, Johannes Lutz and Frahm, Jan-Michael},
    title={Structure-from-Motion Revisited},
    booktitle={Conference on Computer Vision and Pattern Recognition},
    year={2016},
}

@inproceedings{godard2017unsupervised,
  title={Unsupervised monocular depth estimation with left-right consistency},
  author={Godard, Cl{\'e}ment and Mac Aodha, Oisin and Brostow, Gabriel J},
  booktitle={Proceedings of the IEEE conference on computer vision and pattern recognition},
  pages={270--279},
  year={2017}
}

@inproceedings{mildenhall2020nerf,
 title={NeRF: Representing Scenes as Neural Radiance Fields for View Synthesis},
 author={Ben Mildenhall and Pratul P. Srinivasan and Matthew Tancik and Jonathan T. Barron and Ravi Ramamoorthi and Ren Ng},
 year={2020},
 booktitle={ECCV},
}

@article{johnson2017high,
  title={High-resolution underwater robotic vision-based mapping and three-dimensional reconstruction for archaeology},
  author={Johnson-Roberson, Matthew and Bryson, Mitch and Friedman, Ariell and Pizarro, Oscar and Troni, Giancarlo and Ozog, Paul and Henderson, Jon C},
  journal={Journal of Field Robotics},
  volume={34},
  number={4},
  pages={625--643},
  year={2017},
  publisher={Wiley Online Library}
}

@inproceedings{joshi2022underwater,
  title={Underwater exploration and mapping},
  author={Joshi, Bharat and Xanthidis, Marios and Roznere, Monika and Burgdorfer, Nathaniel J and Mordohai, Philippos and Li, Alberto Quattrini and Rekleitis, Ioannis},
  booktitle={2022 IEEE/OES Autonomous Underwater Vehicles Symposium (AUV)},
  pages={1--7},
  year={2022},
  organization={IEEE}
}

@inproceedings{kinsey2006survey,
  title={A survey of underwater vehicle navigation: Recent advances and new challenges},
  author={Kinsey, James C and Eustice, Ryan M and Whitcomb, Louis L},
  booktitle={IFAC conference of manoeuvering and control of marine craft},
  volume={88},
  pages={1--12},
  year={2006},
  organization={Lisbon}
}

@article{leonard2016autonomous,
  title={Autonomous underwater vehicle navigation},
  author={Leonard, John J and Bahr, Alexander},
  journal={Springer handbook of ocean engineering},
  pages={341--358},
  year={2016},
  publisher={Springer}
}

@inproceedings{cabon2025must3r,
  title={Must3r: Multi-view network for stereo 3d reconstruction},
  author={Cabon, Yohann and Stoffl, Lucas and Antsfeld, Leonid and Csurka, Gabriela and Chidlovskii, Boris and Revaud, Jerome and Leroy, Vincent},
  booktitle={Proceedings of the Computer Vision and Pattern Recognition Conference},
  pages={1050--1060},
  year={2025}
}

@article{merveille2024advancements,
  title={Advancements in sensor fusion for underwater SLAM: A review on enhanced navigation and environmental perception},
  author={Merveille, Fomekong Fomekong Rachel and Jia, Baozhu and Xu, Zhizun and Fred, Bissih},
  journal={Sensors (Basel, Switzerland)},
  volume={24},
  number={23},
  pages={7490},
  year={2024}
}

@inproceedings{rahman2019svin2,
  title={Svin2: An underwater slam system using sonar, visual, inertial, and depth sensor},
  author={Rahman, Sharmin and Li, Alberto Quattrini and Rekleitis, Ioannis},
  booktitle={2019 IEEE/RSJ International Conference on Intelligent Robots and Systems (IROS)},
  pages={1861--1868},
  year={2019},
  organization={IEEE}
}

@misc{grupp2017evo,
  title={evo: Python package for the evaluation of odometry and SLAM.},
  author={Grupp, Michael},
  howpublished={\url{https://github.com/MichaelGrupp/evo}},
  year={2017}
}

@misc{gtsam,
  author       = {Frank Dellaert and GTSAM Contributors},
  title        = {borglab/gtsam},
  month        = May,
  year         = 2022,
  publisher    = {Georgia Tech Borg Lab},
  version      = {4.2a8},
  doi          = {10.5281/zenodo.5794541},
  url          = {https://github.com/borglab/gtsam)}}
}

\end{document}